%% file: sample-sigconf.tex
\documentclass[sigconf]{acmart}

\usepackage{tabularx}
\usepackage{amsfonts}
\usepackage{amsmath}
\usepackage{booktabs}
\usepackage{multirow}
\usepackage{algorithm}
\usepackage{algpseudocode}
\usepackage{graphicx}
\usepackage{CJKutf8}
\usepackage[normalem]{ulem}
\usepackage{mathrsfs}
\usepackage{bbm}
\usepackage{arydshln}
\newcommand{\pan}[1]{\textcolor{cyan}{From Pan: #1}}
\newcommand{\nop}[1]{}
\newcommand{\add}[1]{\textcolor{red}{#1}}
\newcommand\BibTeX{B\textsc{ib}\TeX}
\DeclareMathOperator*{\argmax}{arg\,max}
\DeclareMathOperator*{\argmin}{arg\,min}

\AtBeginDocument{%
  \providecommand\BibTeX{{%
    \normalfont B\kern-0.5em{\scshape i\kern-0.25em b}\kern-0.8em\TeX}}}

 \setcopyright{none} 
\renewcommand\footnotetextcopyrightpermission[1]{} 
\makeatletter
\renewcommand\@formatdoi[1]{\ignorespaces}
\makeatother

\acmConference{Arxiv}{Preprint}{Manuscript}

\begin{document}

\title{Towards Evaluating the Robustness of Chinese BERT Classifiers}

\author{Boxin Wang}
\affiliation{%
  \institution{University of Illinois at Urbana-Champaign}
}
\email{boxinw2@illinois.edu}

\author{Boyuan Pan}
\affiliation{%
  \institution{Zhejiang University}
}
\email{panby@zju.edu.cn}

\author{Xin Li}
\affiliation{%
  \institution{Tencent}
}
\email{alonsoli@tencent.com	}

\author{Bo Li}
\affiliation{%
 \institution{University of Illinois at Urbana-Champaign}
}
\email{lbo@illinois.edu}


\input{abstract.tex}




\maketitle

\input{introduction.tex}
\input{related_works.tex}

\input{model.tex}

\input{experiment.tex}

\bibliographystyle{ACM-Reference-Format}
\bibliography{sample-base,acl2020}
\input{appendix.tex}

\end{document}

%% file: abstract.tex
\begin{abstract}
  Recent advances in large-scale language representation models such as BERT have improved the state-of-the-art performances in many NLP tasks. Meanwhile, character-level Chinese NLP models, including BERT for Chinese, have also demonstrated that they can outperform the existing models. In this paper, we show that, however, such BERT-based models are vulnerable under character-level adversarial attacks. We propose a novel Chinese char-level attack method against BERT-based classifiers. Essentially, we generate ``small" perturbation on the character level in the embedding space and guide the character substitution procedure. Extensive experiments show that the classification accuracy on a Chinese news dataset drops from $91.8\%$ to $0\%$ by manipulating less than 2 characters on average based on the proposed attack. Human evaluations also confirm that our generated Chinese adversarial examples barely affect human performance on these NLP tasks.

\end{abstract}

%% file: introduction.tex
\section{Introduction}
\nop{
\nop{Pretrained language representation models, including BERT \citep{Devlin2019BERTPO} and XLNet \citep{Yang2019XLNetGA}, have obtained state-of-the-art results over many downstream NLP tasks.}
\add{Recently, the impressive performance of BERT~\citep{Devlin2019BERTPO} has inspired many pre-trained large-scale language models \citep{Yang2019XLNetGA}~\pan{add more citations}, which have obtained state-of-the-art results over many downstream NLP tasks.}
\citet{Tenney2019BERTRT} points out BERT-based models can disambiguate information from high-level representation, which makes them effective in ambiguous languages 
\sout{ (such as Chinese). In Chinese, there are no explicit word delimiters, and the granularity of words is less well defined than other languages (such as English) \citep{ding-etal-2019-neural}}.
Moreover, \citet{li-etal-2019-word-segmentation} find that character-based models (\textit{e.g.} BERT) outperform word-based models \sout{that} \add{in the Chinese environment since the latter} are vulnerable to data sparsity and out-of-vocabulary \add{situations}\nop{in the Chinese environment}. \add{In Chinese, there are no explicit word delimiters, and the granularity of words is less well defined than other languages (such as English) \citep{ding-etal-2019-neural}.}}

Recently, the impressive performance of BERT~\citep{Devlin2019BERTPO} has inspired many pre-trained large-scale language models \citep{Yang2019XLNetGA, zhang-etal-2019-ernie,Lan2019ALBERTAL}, which have obtained state-of-the-art results over many downstream NLP tasks. Besides its dominant performance in English datasets, \citet{Tenney2019BERTRT} point out that BERT is also effective in ambiguous languages such as Chinese, whose granularity of words is less well defined than English \citep{ding-etal-2019-neural}, because BERT models can disambiguate information from high-level representation.
Moreover, \citet{li-etal-2019-word-segmentation} find that in the Chinese environment, using character-based models (\textit{e.g.,} BERT) is more suitable than word-based models, as the latter often suffer from data sparsity and out-of-vocabulary problems. 


However, are Chinese char-based models such as BERT robust under adversarial settings? To the best of our knowledge, we are the first to study this problem in the Chinese domain. While a large number of studies focus on generating adversarial examples in the continuous data domain (e.g. image and audio), generating adversarial text examples in the discrete domain is much more challenging. Current adversarial text generation work  \citep{jia-liang-2017-adversarial, Jin2019IsBR, Li2018TextBuggerGA,alzantot-etal-2018-generating} is mainly heuristic and not scalable to Chinese in that Chinese characters are intrinsically polysemous. Some char-level adversarial attacks in the English context \citep{ebrahimi-etal-2018-hotflip} are shown ineffective for Chinese char-level attacks, as the sizes of candidate characters increase by two orders of magnitude and the computational costs surge, especially for BERT-based classifiers.

\begin{table}[t]\small \setlength{\tabcolsep}{7pt}
\centering
\begin{CJK*}{UTF8}{gbsn}
\begin{tabular}{p{7.8cm}}
\toprule 
\textbf{Original Chinese Text: } 5 名少年抢劫超市杀害\textcolor{blue}{女}老板被刑拘 \\
(\textbf{Translation:} 5 youngsters were arrested for robbing the convenience store and killing the \textcolor{blue}{female} owner. ) \\
\textbf{Adversarial Chinese Text: } 5 名少年抢劫超市杀害\textcolor{red}{庾}老板被刑拘
 \\
(\textbf{Translation: } 5 youngsters were arrested for robbing the convenience store and killing the owner \textcolor{red}{Yu}. )\\ \hdashline[1pt/2pt]
\textbf{Topic Prediction: }  \textcolor{blue}{Society News} $\rightarrow$  \textcolor{red}{Entertainment News} \\
\midrule
\textbf{Original Chinese Text: } 网 富 \textcolor{blue}{宝}：网 富 宝 是 一 个 帮 助 用 户 模 拟 和 分 析 基 金 投 资 的 在 线 平 台 ， 采 用 交 互 式 可 视 化 方 法 展 现 中 国 基 金 市 场 \\
(\textbf{Translation:} 
Wangfu\textcolor{blue}{bao}: Wangfubao is an on-line platform that helps users to simulate and analyze the investment of funds. It uses interactive visual methods to display the fund market in China. ) \\
\textbf{Adversarial Chinese Text: } 网 富 \textcolor{red}{气}：网 富 宝 是 一 个 帮 助 用 户 模 拟 和 分 析 基 金 投 资 的 在 线 平 台 ， 采 用 交 互 式 可 视 化 方 法 展 现 中 国 基 金 市 场 
 \\
(\textbf{Translation: } Wangfu\textcolor{red}{qi}: Wangfubao is an on-line platform that helps users to simulate and analyze the investment of funds. It uses interactive visual methods to display the fund market in China.)  \\  \hdashline[1pt/2pt]
\textbf{Account Prediction: } \textcolor{blue}{Funds} $\rightarrow$  \textcolor{red}{Other Financial Managements} \\
\bottomrule
\end{tabular}
\end{CJK*}
\caption{{\small Two adversarial examples generated by \texttt{AdvChar} for Chinese BERT classifiers on the THUCTC and Wechat Finance datasets. Simply replacing \textcolor{blue}{one character} with \textcolor{red}{another} can lead the \textcolor{blue}{correct prediction} to a \textcolor{red}{wrong one}.}}
\label{fig:qa_pipeline}
\vspace{-10mm}
\end{table}

In this paper, we propose an efficient Chinese char-level adversarial attack approach \texttt{AdvChar} towards evaluating the robustness of BERT-based models. Our algorithm adds perturbation in the character embedding space guided by the gradient and controls the magnitude of perturbation as small as possible so that only a very few characters are replaced. 
We find out that existing char-level Chinese BERT classifiers are vulnerable under our adversarial attack. Model accuracy can drop from $91.8\%$ to $0\%$ by simply flipping one or two characters, as Figure \ref{fig:qa_pipeline} shows. Though these adversarial examples are generated in the whitebox settings, they can transfer to blackbox classifiers and significantly impair their performances. 

In summary, our contributions lie on: 

1) To the best of our knowledge, this is the first study to generate Chinese adversarial examples against the state-of-the-art BERT classifiers. 

2) We propose a novel and efficient Chinese char-level adversarial targeted and untargeted attack algorithm that is able to handle thousands of Chinese characters and perturb a very few of tokens.  Both quantitative and human evaluations demonstrate the effectiveness and validity of our attacks over BERT-based models on several large-scale Chinese datasets. 


%% file: related_works.tex
\section{Related Works}

In contrast to a large amount of adversarial attacks in the continuous data domain  \citep{Yang2018CharacterizingAA,Carlini2018AudioAE,Eykholt2017RobustPA}, there are a few studies focusing on the discrete text domain. 
\citet{jia-liang-2017-adversarial} use handcrafted rule-based heuristic methods along with crowd-sourcing to generate valid adversarial sentences and fool the question answering models. Similarly, \citet{Thorne2019AdversarialAA,niven-kao-2019-probing} use rule-based heuristics to attack specific tasks such as fact verification and argument reasoning comprehension. However, these methods cannot be applied to other NLP tasks~\cite{pan2018macnet,pan2019reinforced} and need human-crafted rules to guide the search. To automatically find the valid adversarial examples, \citet{ebrahimi-etal-2018-hotflip} propose a whitebox gradient-based attack method to find character-level substitution. However, this method is not efficient when it comes to the Chinese language, where there are thousands of common characters compared with 26 English characters. Additionally, the following studies \citep{alzantot-etal-2018-generating, ren-etal-2019-generating, zhang-etal-2019-generating-fluent, Jin2019IsBR} consider different strategies to perform word-level replacement while maintaining grammatical correctness and sematic similarity.

In addition, while pretrained language models such as BERT \citep{Devlin2019BERTPO} and XLNet \cite{Yang2019XLNetGA} have achieved state-of-the-art results in various NLP tasks, the robustness of these language models are challenged. \citet{niven-kao-2019-probing} point out that BERT is only learning the statistical cues, which can be broken by simply putting negation ahead. \citet{Jin2019IsBR} also finds BERT is vulnerable under adversarial attacks. 


%% file: model.tex
\section{Methodology}

\nop{We describe the methodology of our proposed char-level adversarial generation \texttt{AdvChar} in this section. \texttt{AdvChar} is an efficient attack algorithm performed in the whitebox settings, where it has access to all the parameters of the models along with the gradient information. Generally speaking, \texttt{AdvChar} adds perturbation to each Chinese character and uses the gradient of the input to guide the character substitution. In targeted attack and untargeted attack scenarios, we define two loss functions, based on which we optimize the perturbation to achieve the respective goals, and meanwhile control the magnitude of the perturbation. }

\subsection{Problem Formulation}

Given the input $\boldsymbol{x}= [\boldsymbol{x_0, x_1, ..., x_n}]$, where $\boldsymbol{x_0}$ is a special token \texttt{[CLS]} prepended to every input and $\boldsymbol{x_i}$ is a Chinese character, the BERT-based classification model $f$ maps the input to the final logits $\boldsymbol{z} = f(\boldsymbol{x}) \in \mathbb{R}^{C}$, where $C$ is the number of class, and outputs a label $y = \argmax f(\boldsymbol{x})$. Formally, the BERT-based classifier $f$ first encodes the input \nop{The logits $\boldsymbol{z}$ is generated by first encoding the input}
\begin{equation}
    [\boldsymbol{h_0, h_1, ..., h_n}] = \text{BERT}([\boldsymbol{x_0, x_1, ..., x_n})],
\end{equation}
and outputs the logits $\boldsymbol{z}$ via a fully connected layer based on the hidden state $\boldsymbol{h_0}$ of \texttt{[CLS]}, which represents the sentence embedding for classification tasks \citep{Devlin2019BERTPO}. 

During the adversarial evaluation, we investigate our attack algorithm efficiency by calculating the targeted attack success rate (TSR):
\begin{equation}
    \text{TSR} = \frac{1}{|D_{\text{adv}}|} \sum_{\boldsymbol{x'} \in D_{\text{adv}}}{\mathbbm{1}[\argmax f(\boldsymbol{x'}) \equiv {y^*}]}
    \label{eq:tsr}
\end{equation}
and untargeted attack success rate (USR):
\begin{equation}
    \text{USR} = \frac{1}{|D_{\text{adv}}|} \sum_{ \boldsymbol{x'} \in D_{\text{adv}}}{\mathbbm{1}[\argmax f(\boldsymbol{x'}) \neq {y}]}
    \label{eq:usr}
\end{equation}
where $D_{\text{adv}}$ is the adversarial dataset, $y^*$ is the targeted false class, $y$ is the ground truth label, and $\mathbbm{1}(\cdot)$ is the indicator function.
\subsection{Algorithm}
The whole pipeline is shown in Algorithm \ref{algo}.

\textbf{Character Substitution Procedure.} Due to the discrete nature of text, it is hard to directly utilize the gradient to guide character substitution in the character space. However, in BERT, each discrete character $\boldsymbol{x_i} \in \mathbb{R}^{|V|}$ (one-hot vector, where $V$ is the Chinese character set) is mapped into a high-dimensional embedding space of dimension $d_c$ via the BERT embedding matrix $\boldsymbol{M_e} \in \mathbb{R}^{d_c\times |V|}$
\begin{equation}
    [\boldsymbol{e_1, e_2, ..., e_n}] = \boldsymbol{M_e}\Big[\boldsymbol{x_0, x_1, ..., x_n}\Big].
\end{equation}.

Therefore, we can search the perturbation in the embedding space and map the perturbed character embedding back to characters. Suppose we already have an optimal perturbation $\boldsymbol{e^*}$ in the embedding space that can achieve the attack goal and is the minimal perturbation. We can choose the perturbed character $\boldsymbol{x'_i}$ as the semantically closest character to the perturbed embedding $\boldsymbol{e'_i}$ 
\begin{align}
    & \boldsymbol{e'_i} = \boldsymbol{e_i} + \boldsymbol{e^*_i}, \\
    & \boldsymbol{x'_i} = \argmin ( [\boldsymbol{e'_i; e'_i; ..., e'_i}] - \boldsymbol{M_e}).
\end{align}

 If we control the perturbation $\boldsymbol{e^*}$ to be small enough, most characters will remain the same and a very few characters is perturbed to its semantic close neighbors. In this way, the adversarial examples look still valid to the human but can fool the machines.

\begin{algorithm}[t]
  \caption{\texttt{AdvChar}: Generating Chinese Character-level adversarial examples} \label{algo}
  \begin{flushleft} 
  \textbf{Input:} Chinese characters $\boldsymbol{x} = [\boldsymbol{x_0, x_1, ..., x_n}]$, BERT-based classifier $f:\boldsymbol{x} \rightarrow \boldsymbol{z}$ maps input to logits, objective attack function $g(\cdot)$,  embedding matrix $\boldsymbol{M_e}$, constants $c$ and $\kappa$, max steps $m$ \\
  \textbf{Output:} Adversarial Chinese characters $x'$
  \end{flushleft} 
  \begin{algorithmic}[1]
  \State Initialize perturbation $\boldsymbol{e^*_{{}_{0}}} \leftarrow 0$
  \State $\boldsymbol{e} \leftarrow \boldsymbol{M_e} \boldsymbol{x}$
  \For{$k=0,1,...,m-1$}
    \State $\boldsymbol{e'} \leftarrow \boldsymbol{e} + \boldsymbol{e^*_{{}_{k}}}$
    \State \emph{// Phase I: Character Substitution}
    \For{$i = 1,2,...,n$} 
        \State $ \boldsymbol{x'_i} \leftarrow \argmin ( [\boldsymbol{e'_i; e'_i; ..., e'_i}] - \boldsymbol{M_e})$
    \EndFor
    \State \emph{// Phase II: Optimize over the $\boldsymbol{e^*_{{}_{k}}}$}
    \State$ \mathcal{L}(\boldsymbol{e^*_{{}_{k}}}) \leftarrow  ||\boldsymbol{e^*_{{}_{k}}}||_p + c \cdot g(\boldsymbol{x'})$
    \State $\boldsymbol{e^*_{{}_{k+1}}} \leftarrow \boldsymbol{e^*_{{}_{k}}} - \alpha \nabla \mathcal{L}(\boldsymbol{e^*_{{}_{k}}})$
  \EndFor
  \State \textbf{return} $\boldsymbol{x'}$
  \end{algorithmic}
\end{algorithm}

\begin{table*}[t]\small \setlength{\tabcolsep}{7pt}
\centering
\begin{tabular}{cccccccccc}
\toprule
\multirow{2}{*}{Dataset} & Original & & \multicolumn{3}{c}{AdvChar (untargeted)} & \multicolumn{3}{c}{AdvChar (targeted)} & Baseline\\
\cmidrule(lr){4-6} \cmidrule(lr){7-9}  \cmidrule(lr){10-10}
& Acc & $c/\kappa$ & $5/5$ & $10/5$  & $10/10$ & $5/5$ & $10/5$  & $10/10$ & (untargeted) \\
\midrule
\multirow{3}{*}{THUCTC} & \multirow{3}{*}{0.918} & TSR & -  & - & - & 0.941 & \textbf{0.945}  & \textbf{0.945}    & -    \\
    &  & USR & \textbf{1.000}          & \textbf{1.000}  & \textbf{1.000}  & 0.953 &  \textbf{0.958} &  \textbf{0.958} & 0.040\\
    & & edit distance & \textbf{1.583}         & 1.690   & 1.718 & \textbf{2.924} &  3.186 & 3.045 & 2.000 \\
\midrule
\multirow{3}{*}{Wechat} & \multirow{3}{*}{0.886} & TSR & -  & -   & -  & 1.000 & 0.999 & 0.999 & -  \\
     & & USR & \textbf{1.000}         & \textbf{1.000}          & \textbf{1.000}  & \textbf{1.000} & 0.999 & 0.999 & 0.006   \\
     & & edit distance & \textbf{2.573}  & 2.599   & 4.282 & 6.364 &  6.376 & 12.58 & 3.000\\
\bottomrule
\end{tabular}
\caption{Attack success rates on Chinese BERT-based classifier for two datasets. 
} \label{WhiteboxSentiment}
\vspace{-6mm}
\end{table*}

\textbf{Optimization-based Search.}
We use the neural network to search for the optimal perturbation variable $\boldsymbol{e^*}$. We freeze all the parameters of the BERT-based classifier $f$ and optimize the only variable $\boldsymbol{e^*}$. Following \citet{Carlini2016TowardsET}, we define the loss function as

\begin{equation}
    \mathcal{L}(\boldsymbol{e^*})  =  ||\boldsymbol{e^*}||_p + c \cdot g(\boldsymbol{x'}),
    \label{cw}
\end{equation}
where the first term 
controls the magnitude of perturbation, while $g(\cdot)$ is the attack objective function depending on the attack scenario. $c$ weighs the attack goal against the attack cost. 

In the \emph{targeted attack} scenario, we define $g(\cdot)$ as
\begin{equation*}
    g(\boldsymbol{x'}) = \text{max}[\text{max}\{f(\boldsymbol{x'})_i:i \neq t\} - f(\boldsymbol{x'})_t, -\kappa],
\end{equation*}
where $t$ is the targeted false class and $f(\boldsymbol{x'})_i$ is the $i$-th class logit. A larger $\kappa$ encourages the classifier output targeted false class with higher confidence. 

In the \emph{untargeted attack} scenario, $g(\cdot)$ becomes
\begin{equation*}
    g(\boldsymbol{x'}) = \text{max}[f(\boldsymbol{x'})_t - \text{max}\{f(\boldsymbol{x'})_i:i \neq t\} , -\kappa],
\end{equation*}
where $t$ is the ground truth class.

%% file: experiment.tex
\section{Experimental Results}
\label{sec:exp}


We conduct experiments on two Chinese classification datasets. We first perform the adversarial evaluation in the whitebox settings and validate the effectiveness of our proposed attack. We also explore the transferability of these adversarial examples. The following human evaluation confirms that our generated adversarial examples barely affect human performances. 

\subsection{Datasets}

\indent \textbf{THUCTC \citep{sun2016thuctc}} is a public Chinese news classification dataset. It consists of more than 740k news articles between 2005 and 2011  extracted from Sina News. These articles are classified into 14 categories, including education, technology, society and politics. To speed up the evaluation process, we use the news titles for the classification. We evenly sampled articles from all classes. We use 585,390 articles as the training set, 250,682 articles as the development set, and another 1,000 articles as the testing set to perform the adversarial evaluation. 

\textbf{Wechat Finance Dataset.} This dataset is a private dataset from the Wechat team, who collect 13,051 subscription accounts in the finance domain. Based on the account description, they use crowd-sourcing to classify the account into 11 sub-classes, including insurance, banks, credit cards and funds. Each account description has 94.18 Chinese characters on average. We split the dataset into the training set (10,000 descriptions), the validation set (1,163 descriptions) and the rest as the testing set (1,888 descriptions).


\subsection{Adversarial Evaluation}
\textbf{Baseline.}
As there are no existing efficient Chinese character-level adversarial approaches, we propose a simple attack strategy as our baseline. We first cluster the character embedding by K-means and generate 1,000 embedding clusters. During attack, we randomly choose two or three characters and replace each of them with a random character belong to another random cluster. 

\textbf{Implementation Details.}
We set the max optimization steps $m$ to 100 and use $\ell_2$ norm in the loss function (equation \ref{cw}) that is iteratively optimized via Adam \citep{Kingma2014AdamAM}. We use the pretrained BERT-base model and fine-tune BERT on each dataset independently with a batch size of 64, learning rate of 2e-5 and early stopping. We experiment with different attack strategies, \textit{e.g.,} setting the targeted false class as a specific class or the numerically next class to the ground truth. In this paper, we choose the targeted attack class as ``entertainment news'' for THUCTC dataset and ``Fund Account'' for Wechat Finance Dataset (when the ground truth label is the targeted class, we switch the target to another random class), which respectively achieves the highest targeted attack success rate in a set of ablation studies . More detailed settings of BERT-based classifiers and adversarial analysis on adversarial attack success rates are discussed in Appendix \ref{appdix:model} and \ref{appdix:analysis}.

\begin{table}[t]\small \setlength{\tabcolsep}{7pt}
\centering
\begin{tabular}{ccc}
\toprule
Dataset & AdvChar (target) & AdvChar (untarget) \\
\midrule
THUCTC & \textbf{0.491} & 0.569  \\
Wechat & \textbf{0.143} & 0.552  \\
\bottomrule
\end{tabular}
\caption{Classifier accuracy after transerability-based blackbox attack on other BERT-based classifiers.} \label{blackbox}
\vspace{-10mm}
\end{table}

\textbf{Results.} 
We perform our char-level adversarial attack on BERT-based classifiers for two datasets in both targeted and untargeted attack scenarios. The attack results are shown in Table \ref{WhiteboxSentiment}. 

We can see the untargeted attack can always achieve $100\%$ attack success rate on both datasets, making the model performance drop to $0\%$ by manipulating merely less than two tokens on average on the Chinese News Dataset. In the targeted attack scenario, we can always make BERT output our expected false class on the Wechat dataset, and achieve around $95\%$ targeted attack success rate on the THUCTC dataset. 

Surprised by the fragility of Chinese BERT, we conduct several case studies on the generated adversarial text. We conjecture that Chinese BERT classifiers tend to make predictions based on a certain set of characters (statistical cues) without understanding the sentences. Therefore, the adversarial attack can easily succeed by replacing such critical characters. For the topic prediction example in Table \ref{fig:qa_pipeline}, ``Yu'' is a Chinese celebrity name  and only appears in the Entertainment News in the training set. Therefore, the BERT classifier takes ``Yu'' as a strong signal to classify the news as the Entertainment News. Similarly, another wrong account prediction in Table \ref{fig:qa_pipeline} is because term ``qi'' is a frequent financial products (petroleum and gas) used in other financial management accounts. 

We also find that increasing the constants $c$ and $\kappa$ can improve the attack success rate at the cost of more perturbed characters. Additionally, because the Wechat dataset has longer text than the THUCTC dataset, it is expected that more characters are manipulated on the Wechat dataset in order to break the statistical cues.

\textbf{Transferability.} 
The previous experiments are conducted in the whitebox settings, \textit{i.e.,} we have all the access to the model parameters and gradients, which is a strong assumption that does not often hold in the real world. Therefore, we want to further investigate whether our algorithm can still attack BERT classifiers in the blackbox settings, when we cannot access model parameters. In this setting, we can attack a blackbox BERT classifier of different parameters by using the adversarial text generated from a whitebox BERT trained by ourselves. 
 

The transferability-based attack results are shown in Table \ref{blackbox}.
We find the adversarial text can still substantially affect the accuracy of blackbox models. In addition, the targeted attack success rate turns out to be stronger than the untargeted attack. Particularly, the targeted adversarial examples on the Wechat dataset can make the blackbox BERT classifier performance drop from $88.2\%$ to $14.3\%$.

\begin{table}[t]\small \setlength{\tabcolsep}{7pt}
\centering
\begin{tabular}{ccc}
\toprule
Dataset & Clean & Adversarial  \\
\midrule
Human & 0.84 $\pm 0.04$ &  0.80 $\pm 0.06$   \\
BERT  & 0.82 & 0.00  \\
\bottomrule
\end{tabular}
\caption{Human performances compared with BERT classifiers in original dataset and adversarial dataset. }\label{human}
\vspace{-10mm}
\end{table}

\subsection{Human Evaluation}
To confirm that our generated adversarial examples are valid to human, we conduct the following human evaluation. We randomly sample 50 clean sentences and 50 adversarial sentences generated by \texttt{AdvChar} (untargeted $c/\kappa=5/5$) on the Wechat dataset. We give volunteers two labels: a ground truth label and a fake label. For the clean sentence, the fake label is a random label different from the groud truth. As for the adversarial data, the fake label is the model's wrong prediction.
Both clean text and adversarial text are mixed together. Ten native Chinese student volunteers are asked to choose the correct label. These native Chinese students are not required to be equipped with professional finance knowledge, so their evaluation results could contain errors and may be not as accurate as the financial annotators employed by the Wechat team. 

The evaluation results are shown in Table \ref{human}. We find that our adversarial text barely impacts the human perception, since the human performance on adversarial data is only $4\%$ lower than the clean data, in contrast to the huge performance drop of BERT classifiers from $82\%$ to $0\%$. 

\section{Conclusion}

In this paper, we propose a novel character-level adversarial attack method to probe the robustness of BERT-based Chinese classifiers. Our experiments show that existing character-level BERT-based models are not robust in both whitebox and blackbox settings. While we observe the impressive improvements using the pretrained language models, we expect our study can encourage further research into the robustness problems of current pretrained language understanding models.

%% file: appendix.tex
\onecolumn
\appendix
\section{Appendices}
\label{sec:appendix}

\subsection{Model Settings}
\paragraph{Classifier}
\label{appdix:model}
We use BERT \citep{Devlin2019BERTPO} as the classifier for both datasets. BERT is a transformer \cite{Vaswani2017AttentionIA} based model, which is unsupervisedly pretrained on large Chinese corpuses and is effective for downstream Chinese NLP tasks. We use the 12-layer BERT-base model with 768 hidden units, 12 self-attention heads and 110M parameters. We fine-tune BERT on each dataset independently with a batch size of 64, learning rate of 2e-5 and early stopping.

\subsection{Analysis}
\label{appdix:analysis}
In this section, we will evaluate the possible factors that will affect the attack success rate.

\paragraph{Norm selection.} 
In the main experiment, we use $l_2$ norm for our attack loss function (equation 7). However, because $l_1$ norm is known for good at feature selection and generating sparse features, we conduct the following experiments by set $l_p$ to $l_1$ and make an comparison with $l_2$ norm. The experimental results are shown in Table \ref{l1norm-untargeted} and \ref{l1norm-targeted}. We find the overall attack success rates decrease when switching to $l_1$ norm. However, given the same set of constants $c$ and $\kappa$, we find the $l_1$ attack does change less words.

\begin{table*}[htp!]\small \setlength{\tabcolsep}{7pt}
\centering
\caption{Untargeted attack success rates on Chinese BERT-based classifier for THUCTC dataset. ``target'' and ``untarget'' calculate the targeted attack success rate (equation \ref{eq:tsr}) and the untargeted attack success rate (equation \ref{eq:usr}). ``\#/chars'' counts the number charcters are modified in average.
}
 \label{l1norm-untargeted}
\begin{tabular}{cccccccccc}
\toprule
\multirow{2}{*}{Dataset} & Original & & \multicolumn{3}{c}{AdvChar ($l_2$ untargeted)} & \multicolumn{3}{c}{AdvChar ($l_1$ untargeted)} & Baseline\\
\cmidrule(lr){4-6} \cmidrule(lr){7-9}  \cmidrule(lr){10-10}
& Acc & $c/k$ & $5/5$ & $10/5$  & $10/10$ & $10/10$ & $10/100$  & $20/20$ & (untargeted) \\
\midrule
\multirow{3}{*}{THUCTC} & \multirow{3}{*}{0.918} & target & -  & - & - & - & -  & -  & -    \\
    &  & untarget & \textbf{1.000}          & \textbf{1.000}  & \textbf{1.000}  & 0.983 &  0.983 & \textbf{0.995}  & 0.040\\
    & & \#/chars & \textbf{1.583}         & 1.690   & 1.718 & \textbf{1.577} &  1.614 & 1.884 & 2.000 \\
\bottomrule
\end{tabular}
\vspace{-3mm}
\end{table*}

\begin{table*}[htp!]\small \setlength{\tabcolsep}{7pt}
\centering
\caption{Targeted attack success rates on Chinese BERT-based classifier for THUCTC dataset. ``target'' and ``untarget'' calculate the targeted attack success rate (equation \ref{eq:tsr}) and the untargeted attack success rate (equation \ref{eq:usr}). ``\#/chars'' counts the number charcters are modified in average.
}
 \label{l1norm-targeted}
\begin{tabular}{cccccccccc}
\toprule
\multirow{2}{*}{Dataset} & Original & & \multicolumn{3}{c}{AdvChar ($l_1$ targeted)} & \multicolumn{3}{c}{AdvChar ($l_2$ targeted)} & Baseline\\
\cmidrule(lr){4-6} \cmidrule(lr){7-9}  \cmidrule(lr){10-10}
& Acc & $c/k$ & $10/10$ & $10/20$  & $30/30$ & $5/5$ & $10/5$  & $10/10$ & (untargeted) \\
\midrule
\multirow{3}{*}{THUCTC} & \multirow{3}{*}{0.918} & target & 0.797  & 0.797 &\textbf{ 0.898 }& 0.941 & \textbf{0.945}  & \textbf{0.945}    & -    \\
    &  & untarget & 0.828          & 0.828 & \textbf{0.920}  & 0.953 &  \textbf{0.958} &  \textbf{0.958} & 0.040\\
    & & \#/chars & 2.000 & \textbf{1.956}   & 3.280 & \textbf{2.924} &  3.186 & 3.045 & 2.000 \\
\bottomrule
\end{tabular}
\vspace{-3mm}
\end{table*}

\paragraph{Attack Strategy.}
As we have achieved $100\%$ attack success rate in the untargeted attack scenario, we now focus on the targeted attack scenario and see which factor contributes to the targeted attack success rate. It is straightfoward to think different targeted attack strategy will impact the targeted attack success rate, because maybe some classes look "farther" than semantic closer classes. So we tried two strategies on THUCTC dataset: 1) as used in the main paper, we set the targeted false class as ``entertainment news''. 2) we enumerate all the classes and set the targeted false class to be numerically the next class. The targeted attack success rate is shown in Table \ref{strategy}. We do find choosing different attack strategy will impact the attack success rate.

\begin{table*}[htp!]\small \setlength{\tabcolsep}{7pt}
\centering
\caption{Attack success rates on Chinese BERT-based classifier for two datasets. ``target'' and ``untarget'' calculate the targeted attack success rate (equation \ref{eq:tsr}) and the untargeted attack success rate (equation \ref{eq:usr}). ``\#/chars'' counts the number charcters are modified in average.
}
 \label{strategy}
\begin{tabular}{cccccc}
\toprule
\multirow{2}{*}{Dataset} & Original & & \multicolumn{2}{c}{AdvChar (targeted $c/\kappa=10/10$)} & Baseline\\
\cmidrule(lr){4-5}  \cmidrule(lr){6-6}
& Acc &  &  strategy 1   &  strategy 2 & (untargeted) \\
\midrule
\multirow{3}{*}{THUCTC} & \multirow{3}{*}{0.918} & target &  \textbf{0.945} & 0.903    & -    \\
    &  & untarget  &  \textbf{0.958} & 0.913 & 0.040\\
    & & \#/chars  & 3.045 & 4.543 & 2.000 \\
\bottomrule
\end{tabular}
\vspace{-3mm}
\end{table*}

\newpage

\subsection{Chineses Adversarial Examples}
\label{appendix:examples}
\begin{CJK*}{UTF8}{gbsn}
\begin{table*}[htp!]\small \setlength{\tabcolsep}{7pt}
\centering
\caption{Chinese Adversarial Examples generated by \texttt{AdvChar} for BERT-based classifier.}
 \label{ansqasentexamples}
\begin{tabular}{p{13.8cm}}
\toprule Input (\textcolor{red}{red} = Modified character, \textbf{bold}=original character.) \\
\midrule

\textbf{Original Chinese Text: } 乌 鲁 木 齐 中 考 答 案 超 出 答 题 边 \textbf{框}视 为 无 效 \\
\textbf{Translation:} Wu Lu Mu Qi's Middle School Examination does not take into consideration answers exceeds the border \textbf{box}.  \\
\\
\textbf{Adversarial Chinese Text: } 乌 鲁 木 齐 中 考 答 案 超 出 答 题 边 \textcolor{red}{躯}视 为 无 效 \\
\textbf{Translation: } Wu Lu Mu Qi's Middle School Examination does not take into consideration answers exceeds the border \textcolor{red}{body}. \\
\textbf{Model Prediction: } Education News （教育新闻） $\rightarrow$ Society News （社会新闻）\\
\\
\midrule

\textbf{Original Chinese Text: } \textbf{教}授 称 出 租 车 特 许 经 营 无 法 可 依 提 出 审 查 被 驳 \\
\textbf{Translation:} \textbf{A professor} said that there was no official laws for the cab drivers to legally run business and proposed to review the laws but refuted. \\
 \\
\textbf{Adversarial Chinese Text: } \textcolor{red}{美}授 称 出 租 车 特 许 经 营 无 法 可 依 提 出 审 查 被 驳
 \\
\textbf{Translation: } \textcolor{red}{An American professor} said that there was no official laws for the cab drivers to legally run business and proposed to review the laws but refuted. \\
\textbf{Model Prediction: } Society News （社会新闻）$\rightarrow$ Stock News （股票新闻） \\
\\
\midrule

\textbf{Original Chinese Text: } 15 \textbf{比}7 ！ 专 家 一 边 倒 支 持 热 火 美 球 迷 却 挺 小 牛 \\
\textbf{Translation:} 15 \textbf{to} 7! Sports Professors all support Miami Heat but American fans vote for Dallas Mavericks  \\
 \\
\textbf{Adversarial Chinese Text: } 15\textcolor{red}{68} 7 ！ 专 家 一 边 倒 支 持 热 火 美 球 迷 却 挺 小 牛
 \\
\textbf{Translation: } 15\textcolor{red}{68} 7! Sports Professors all support Miami Heat but American fans vote for Dallas Mavericks \\
\textbf{Model Prediction: } Sports News （体育新闻） $\rightarrow$ Lottery News（彩票新闻） \\
 \\
\midrule

\textbf{Original Chinese Text: } \textbf{教}师 要 求 表 现 差 学 生 交 2000 元 入 学 保 证 金  \\
\textbf{Translation:} The \textbf{teacher} asked the students who showed poor performance to pay 2,000 yuan for the enrollment certificate  \\
 \\
\textbf{Adversarial Chinese Text: } \textcolor{red}{蔚}师 要 求 表 现 差 学 生 交 2000 元 入 学 保 证 金
 \\
\textbf{Translation: } The  \textcolor{red}{teacher Wei} asked the students who showed poor performance to pay 2,000 yuan for the enrollment certificate \\
\textbf{Model Prediction: } Society News （社会新闻）$\rightarrow$  Education News （教育新闻）\\
 \\
\midrule

\textbf{Original Chinese Text: } 贝 \textbf{卢}斯 科 尼 达 成 离 婚 协 议 每 月 付 35 万 欧 元 赡 养 费  \\
\textbf{Translation:} \textbf{Be\underline{ru}skoenida} Reach a divorce agreement to Pay 350,000 Euros a Month for Maintenance  \\
 \\
\textbf{Adversarial Chinese Text: }  贝 \textcolor{red}{民}斯 科 尼 达 成 离 婚 协 议 每 月 付 35 万 欧 元 赡 养 费
 \\
\textbf{Translation: } \textcolor{red}{Be\underline{min}skoenida} Reach a divorce agreement to Pay 350,000 Euros a Month for Maintenance \\
\textbf{Model Prediction: } Politics News （时政新闻） $\rightarrow$ Society News （社会新闻） \\
 \\
\midrule

\textbf{Original Chinese Text: } 《 \textbf{快}乐 星 球 》 2 月 11 日 登 陆 央 视 将 参 与 网 络 春 晚  \\
\textbf{Translation:} 《\textbf{Happy} Planet》 will land on CCTV on February 11 to participate in the Internet Spring Evening.  \\
 \\
\textbf{Adversarial Chinese Text: }  《 \textcolor{red}{末}乐 星 球 》 2 月 11 日 登 陆 央 视 将 参 与 网 络 春 晚
 \\
\textbf{Translation: } 《\textcolor{red}{Final Happy} Planet》 will land on CCTV on February 11 to participate in the Internet Spring Evening.  \\
\textbf{Model Prediction: } Entertainment News （娱乐新闻） $\rightarrow$  Technology News （科技新闻）\\
 \\
\midrule

\textbf{Original Chinese Text: } 债 券 已 处 牛 市 前 夜 \textbf{债} 市 翻 身 6 年 来 最 惨 一 跌 结 束  \\
\textbf{Translation:} Bonds have been in the bull market since yesterday when the \textbf{bond} market goes up and ends the worst drop in six years  \\
 \\
\textbf{Adversarial Chinese Text: } 债 券 已 处 牛 市 前 夜 \textcolor{red}{港} 市 翻 身 6 年 来 最 惨 一 跌 结 束
 \\
\textbf{Translation: } Bonds have been in the bull market since yesterday when the \textcolor{red}{Hongkong} market goes up and ends the worst drop in six years  \\
\textbf{Model Prediction: }  Financial and economic news （财经新闻） $\rightarrow$  Stock news （股票新闻）\\
 \\

\bottomrule
\end{tabular}
\end{table*}
\end{CJK*}